\documentclass{article}
\usepackage{ijcai20}

\usepackage{times}

\usepackage{soul}
\usepackage{url}
\usepackage[hidelinks]{hyperref}
\usepackage[utf8]{inputenc}
\usepackage[small]{caption}
\usepackage{graphicx}
\usepackage{amsmath}
\usepackage{booktabs}
\usepackage{arydshln}
\usepackage{longtable}
\usepackage{xcolor}
\usepackage{amssymb}
\usepackage{amsmath}
\usepackage{adjustbox}
\usepackage{subfigure}
\usepackage{blindtext}
\usepackage{enumerate}
\usepackage{bm}
\usepackage{dsfont}
\usepackage{array}
\usepackage{multirow}
\usepackage{multicol}
\usepackage{hhline}
\usepackage{colortbl}

\newcolumntype{L}[1]{>{\raggedright\arraybackslash}p{#1}}
\newcolumntype{C}[1]{>{\centering\arraybackslash}p{#1}}
\newcolumntype{R}[1]{>{\raggedleft\arraybackslash}p{#1}}

\title{Multi-source Domain Adaptation in the Deep Learning Era: \\A Systematic Survey}

\author{
Sicheng Zhao$^1$\and
Bo Li$^1$\and
Colorado Reed$^1$\and
Pengfei Xu$^2$\and
Kurt Keutzer$^1$\\
\affiliations
$^1$University of California, Berkeley, USA\ \ \ \ \ \ $^2$Didi Chuxing, China\\
\emails
schzhao@gmail.com, drluodian@gmail.com, cjrd@cs.berkeley.edu,\\
xupengfeipf@didiglobal.com, keutzer@berkeley.edu
}

\begin{document}

\maketitle

\begin{abstract}
In many practical applications, it is often difficult and expensive to obtain enough large-scale labeled data to train deep neural networks to their full capability. Therefore, transferring the learned knowledge from a separate, labeled source domain to an unlabeled or sparsely labeled target domain becomes an appealing alternative. However, direct transfer often results in significant performance decay due to \emph{domain shift}. Domain adaptation (DA) addresses this problem by minimizing the impact of domain shift between the source and target domains. Multi-source domain adaptation (MDA) is a powerful extension in which the labeled data may be collected from multiple sources with different distributions. Due to the success of DA methods and the prevalence of multi-source data, MDA has attracted increasing attention in both academia and industry. In this survey, we define various MDA strategies and summarize available datasets for evaluation. We also compare modern MDA methods in the deep learning era, including latent space transformation and intermediate domain generation. Finally, we discuss future research directions for MDA.
\end{abstract}

\section{Background and Motivation}
\label{sec:Motivation}

The availability of large-scale labeled training data, such as ImageNet, has enabled deep neural networks (DNNs) to achieve remarkable success in many learning tasks, ranging from computer vision to natural language processing. For example, the classification error of the ``Classification + localization with provided training data'' task in the Large Scale Visual Recognition Challenge has reduced from 0.28 in 2010 to 0.0225 in 2017\footnote{\url{http://image-net.org/challenges/LSVRC/2017}}, outperforming even human classification. However, in many practical applications, obtaining labeled training data is often expensive, time-consuming, or even impossible. For example, in fine-grained recognition, only the experts can provide reliable labels~\cite{gebru2017fine}; in semantic segmentation, it takes about 90 minutes to label each Cityscapes image~\cite{cordts2016cityscapes}; in autonomous driving, it is difficult to label point-wise 3D LiDAR point clouds~\cite{wu2019squeezesegv2}.

\begin{figure}[!t]
\begin{center}
\centering \includegraphics[width=1.0\linewidth]{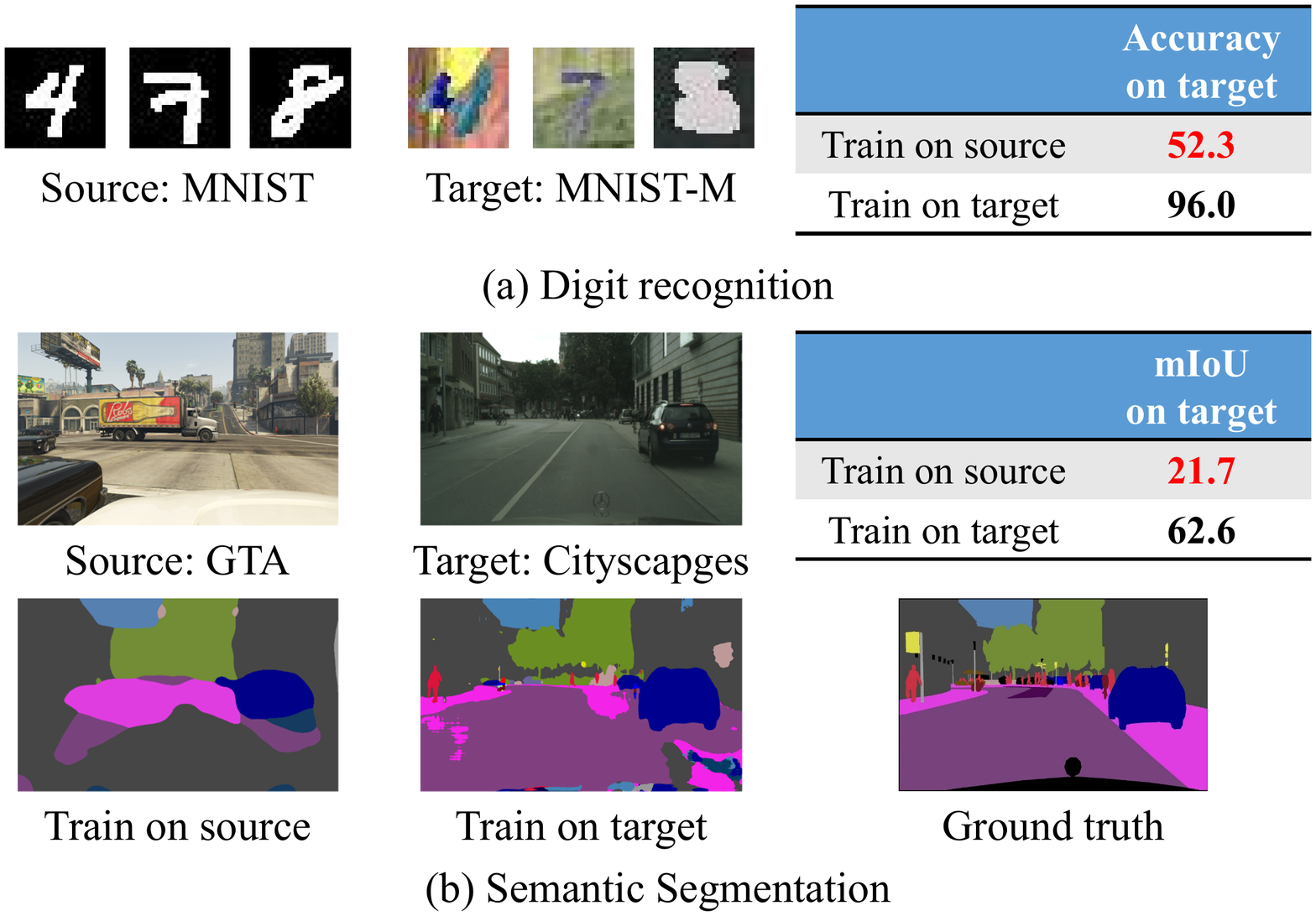}
\caption{An example of \emph{domain shift} in the single-source scenario. The models trained on the labeled source domain do not perform well when directly transferring to the target domain.}
\label{fig:SingleDomainShift}
\end{center}
\end{figure}

One potential solution is to transfer a model trained on a separate, labeled source domain to the desired unlabeled or sparsely labeled target domain. But as Figure~\ref{fig:SingleDomainShift} demonstrates, the \textbf{direct transfer of models across domains leads to poor performance}. Figure~\ref{fig:SingleDomainShift}(a) shows that even for the simple task of digit recognition, training on the MNIST source domain~\cite{lecun1998gradient} for digit classification in the MNIST-M target domain ~\cite{ganin2015unsupervised} leads to a digit classification accuracy decrease from 96.0\% to 52.3\% when training a LeNet-5 model~\cite{lecun1998gradient}. Figure~\ref{fig:SingleDomainShift}(b) shows a more realistic example of training a semantic segmentation model on a synthetic source dataset GTA~\cite{richter2016playing} and conducting pixel-wise segmentation on a real target dataset Cityscapes~\cite{cordts2016cityscapes} using the FCN model~\cite{long2015fully}. If we train on the real data, we obtain a mean intersection-over-union (mIoU) of 62.6\%; but if we train on synthetic data, the mIoU drops significantly to 21.7\%.

\begin{figure}[!t]
\begin{center}
\centering \includegraphics[width=1.0\linewidth]{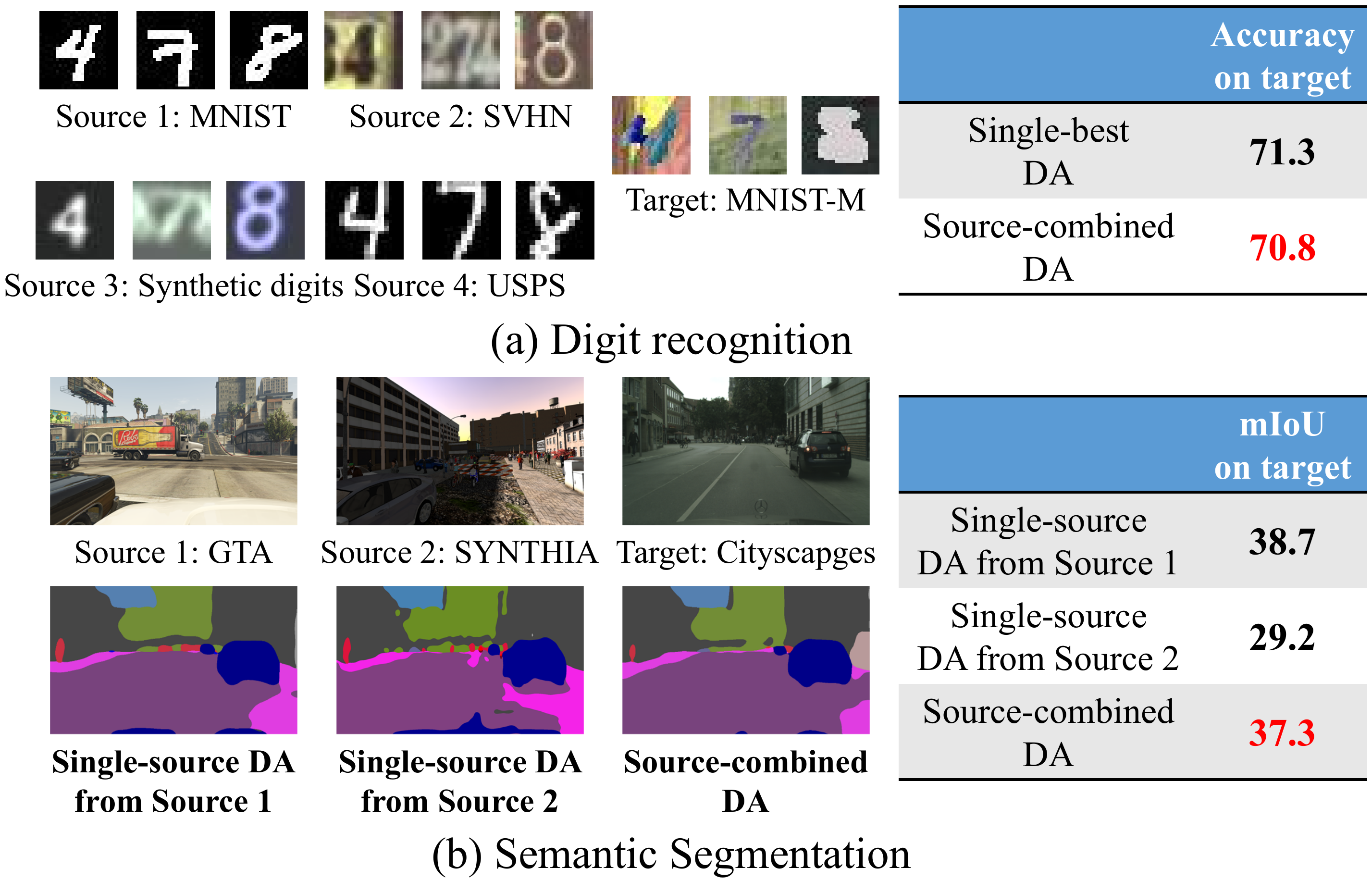}
\caption{An example of \emph{domain shift} in the multi-source scenario. Combining multiple sources into one source and directly performing single-source domain adaptation on the entire dataset does not guarantee better performance compared to just using the best individual source domain.}
\label{fig:MultiDomainShift}
\end{center}
\end{figure}

The poor performance from directly transferring models across domains stems from a phenomenon known as \emph{domain shift}~\cite{torralba2011unbiased,zhao2018emotiongan}: whereby the joint probability distributions of observed data and labels are different in the two domains.
Domain shift exists in many forms, such as from dataset to dataset, from simulation to real-world, from RGB images to depth, and from CAD models to real images.

The phenomenon of domain shift motivates the research on domain adaptation (DA), which aims to learn a model from a labeled source domain that can generalize well to a different, but related, target domain. Existing DA methods mainly focus on the single-source scenario. In the deep learning era, recent single-source DA (SDA) methods usually employ a conjoined architecture with two approaches to respectively represent the models for the source and target domains. One approach aims to learn a task model based on the labeled source data using corresponding task losses, such as cross-entropy loss for classification. The other approach aims to deal with the domain shift by aligning the target and source domains. Based on the alignment strategies, deep SDA methods can be classified into four categories:

\begin{enumerate}
\item \emph{Discrepancy-based methods} try to align the features by explicitly measuring the discrepancy on corresponding activation layers, such as maximum mean discrepancy (MMD)~\cite{long2015learning}, correlation alignment~\cite{sun2017correlation}, and contrastive domain discrepancy~\cite{kang2019contrastive}.
\item \emph{Adversarial generative methods} generate fake data to align the source and target domains at pixel-level based on Generative Adversarial Network (GAN)~\cite{goodfellow2014generative} and its variants, such as CycleGAN~\cite{zhu2017unpaired,zhao2019cycleemotiongan}.
\item \emph{Adversarial discriminative methods} employ an adversarial objective with a domain discriminator to align the features~\cite{tzeng2017adversarial,tsai2018learning}.
\item \emph{Reconstruction based methods} aim to reconstruct the target input from the extracted features using the source task model~\cite{ghifary2016deep}.
\end{enumerate}


In practice, \textbf{the labeled data may be collected from multiple sources with different distributions}~\cite{sun2015survey,bhatt2016multi}. 
In such cases, the aforementioned SDA methods could be trivially applied by combining the sources into a single source: an approach we refer to as \emph{source-combined DA}. However, source-combined DA oftentimes results in a poorer performance than simply using one of the sources and discarding the others. As illustrated in Figure~\ref{fig:MultiDomainShift}, the accuracy on the best single source digit recognition adaptation using DANN~\cite{ganin2016domain} is 71.3\%, while the source-combined accuracy drops to 70.8\%. For segmentation adaptation using CyCADA~\cite{hoffman2018cycada}, the mIoU of source-combined DA (37.3\%) is also lower than that of SDA from GTA (38.7\%). Because the domain shift not only exists between each source and target, but also exists among different sources, the source-combined data from different sources may interfere with each other during the learning process~\cite{riemer2019learning}. Therefore, multi-source domain adaptation (MDA) is needed in order to leverage all of the available data.

The early MDA methods mainly focus on shallow models~\cite{sun2015survey},
either learning a latent feature space for different domains~\cite{sun2011two,duan2012exploiting} or combining pre-learned source classifiers~\cite{schweikert2009empirical}. Recently, the emphasis on MDA has shifted to deep learning architectures. In this paper, we systematically survey recent progress on deep learning based MDA, summarize and compare similarities and differences in the approaches, and discuss potential future research directions.

\section{Problem Definition}
\label{sec:Definition}
In the typical MDA setting, there are multiple source domains $S_1,S_2,\cdots,S_M$ ($M$ is the number of sources) and one target domain $T$. Suppose the observed data and corresponding labels\footnote{The label could be any type, such as object classes, bounding boxes, semantic segmentation, \textit{etc}.} in the $i^{\text{th}}$ source $S_i$ are drawn from distribution $p_i(\mathbf{x}, \mathbf{y})$ are $\textbf{X}_i=\{\textbf{x}_i^j\}_{j=1}^{N_i}$ and $Y_i=\{\mathbf{y}_i^j\}_{j=1}^{N_i}$, respectively, where $N_i$ is the number of source samples. Let $X_T=\{\mathbf{x}_T^j\}_{j=1}^{N_T}$ and $Y_T=\{\mathbf{y}_T^j\}_{j=1}^{N_T}$ denote the target data and corresponding labels drawn from the target distribution $P_T(\textbf{x},\mathbf{y})$, where $N_T$ is the number of target samples.

Suppose the number of labeled target samples is $N_{TL}$, the MDA problem can be classified into different categories:
\begin{itemize}
\item \emph{unsupervised MDA}, when $N_{TL}=0$;
\item \emph{fully supervised MDA}, when $N_{TL}=N_T$;
\item \emph{semi-supervised MDA}, otherwise.
\end{itemize}

\begin{table*}[!t]
\centering\scriptsize
\resizebox{\textwidth}{!}{%
\begin{tabular}
{c c cccccc}
\specialrule{1pt}{1pt}{1pt}
\textbf{Area} & \textbf{Task} & \textbf{Dataset} & \textbf{Reference} & \textbf{\#D} & \textbf{\#S} & \textbf{Labels} & \textbf{Short description}\\
\hline
\multirow{10}{*}{CV}   & \multirow{2}{*}{digit recognition} &  \multirow{2}{*}{Digits-five (D)} & \citeauthor{lecun1998gradient,netzer2011reading} &  \multirow{2}{*}{5}  &  \multirow{2}{*}{145,298}  & \multirow{2}{*}{10 classes} &  \multirow{2}{*}{handwritten, synthetic, and street-image digits}  \\
&  &   & \citeauthor{hull1994database,ganin2015unsupervised} &    &  &  &    \\
\hhline{~|-|-|-|-|-|-|-}
   & \multirow{6}{*}{object classification} &  Office-31 (O) & \citeauthor{saenko2010adapting} &  3  & 4,110 & 31 classes &  images from amazon and taken by different cameras \\
   &  &  Office-Caltech (OC) & \citeauthor{gong2013connecting} &  4  & 2,533 & 10 classes & overlapping categories from Office-31 and \textbf{C}  \\
   &  &  Office-Home (OH) & \citeauthor{venkateswara2017deep} &  4  & 15,500 & 65 classes & artistic, clipart, product, and real objects  \\
   &  &  ImageCLEF (IC) & Challenge\textsuperscript{\ref{ImageCLEF}} &  3  & 1,800 & 12 classes & shared categories from 3 datasets \\
   &  &  PACS (P) & \citeauthor{li2017deeper} &  4  & 9,991 & 7 classes & photographic, artistic, cartoon, and sketchy objects   \\
   &  &  DomainNet (DN)  & \citeauthor{peng2019moment} &  6  & 600,000 & 345 classes &  \tiny{clipart, infographic, artistic, quickdrawn, real, and sketchy objects}  \\
   \hhline{~|-|-|-|-|-|-|-}
   & \multirow{2}{*}{sentiment classification} &  \multirow{2}{*}{SentiImage (SI)} & \citeauthor{machajdik2010affective} &  \multirow{2}{*}{4}  & \multirow{2}{*}{25,986} & \multirow{2}{*}{2 classes} &  \multirow{2}{*}{artistic and social images on visual sentiment}  \\
   &  &  & \citeauthor{you2016building,you2015robust,borth2013large} &   &  &  &    \\
   \hhline{~|-|-|-|-|-|-|-}
   & vehicle counting & WebCamT (W)  & \citeauthor{zhang2017understanding} &  8  & 16,000 & vehicle counts & each camera used as one domain \\
   \hhline{~|-|-|-|-|-|-|-}
   & \multirow{2}{*}{semantic segmentation} &  \multirow{2}{*}{Sim2RealSeg (S2R)} & \citeauthor{cordts2016cityscapes,yu2018bdd100k} &  \multirow{2}{*}{4}  & \multirow{2}{*}{49,366} & \multirow{2}{*}{16 classes} & simulation-to-real adaptation  \\
   &  &   & \citeauthor{richter2016playing,ros2016synthia} &   &  &  & for pixel-wise predictions \\
\hline
\multirow{3}{*}{NLP}  & \multirow{2}{*}{sentiment classification} & AmazonReviews (AR)  & \citeauthor{chen2012marginalized} &  4  & $\approx$12,000  & 2 classes  &  reviews on four kinds of products \\

&  & MediaReviews (MR)  & \citeauthor{liu2017adversarial} &  5  & 6897 & 2 classes & reviews on products and movies \\
\hhline{~|-|-|-|-|-|-|-}

& part-of-speech tagging & SANCL (S)  & \citeauthor{petrov2012overview} & 5 & 5250 & tags & part-of-speech tagging in 5 web domains \\
\specialrule{1pt}{1pt}{1pt}
\end{tabular}
}
\caption{Released and freely available datasets for MDA, where `\#D' and `\#S' represent the number of domains and the total number of samples usually used for MDA, respectively.}
\label{tab:Dataset}
\end{table*}

Suppose $\textbf{x}_i^j\in \mathds{R}^{d_i}$ and $\textbf{x}_T^j\in \mathds{R}^{d_T}$ are an observation in source $S_i$ and target $T$, we can classify MDA into:
\begin{itemize}
\item \emph{homogeneous MDA}, when $d_1=\cdots=d_M=d_T$;
\item \emph{heterogeneous MDA}, otherwise.
\end{itemize}

Suppose $\mathcal{C}_i$ and $\mathcal{C}_T$ are the label set for source $S_i$ and target $T$, we can define different MDA strategies:
\begin{itemize}
\item \emph{closed set MDA}, when $\mathcal{C}_1=\cdots=\mathcal{C}_M=\mathcal{C}_T$;
\item \emph{open set MDA}, for at least one $\mathcal{C}_i$, $\mathcal{C}_i \cap \mathcal{C}_T \subset \mathcal{C}_T$;
\item \emph{partial MDA}, for at least one $\mathcal{C}_i$, $\mathcal{C}_T \subset \mathcal{C}_i$;
\item \emph{universal MDA}, when no prior knowledge of the label sets is available;
\end{itemize}
where $\cap$ and $\subset$ indicate the intersection set and proper subset between two sets.

Suppose the number of labeled source samples is $N_{iL}$ for source $S_i$, the MDA problem can be classified into:
\begin{itemize}
\item \emph{strongly supervised MDA}, when $N_{iL}=N_i$ for $i=1\cdots M$;
\item \emph{weakly supervised MDA},  otherwise.
\end{itemize}

When adapting to multiple target domains simultaneously, the task becomes multi-target MDA. When the target data is unavailable during training~\cite{yue2019domain}, the task is often called multi-source domain generalization or zero-shot MDA.

\section{Datasets}
\label{sec:Datasets}

The datasets for evaluating MDA models usually contain multiple domains with different styles, such as \textit{synthetic} vs. \textit{real}, \textit{artistic} vs. \textit{sketchy}, which impose large domain shift among different domains. Here we summarize the commonly employed datasets in both computer vision (CV) and natural language processing (NLP) areas, as shown in Table~\ref{tab:Dataset}.

\textbf{Digit recognition.} Digits-five includes 5 digit image datasets sampled from different domains, including \emph{handwritten} MNIST (\textbf{mt})~\cite{lecun1998gradient}, \emph{combined} MNIST-M (\textbf{mm})~\cite{ganin2015unsupervised} from MNIST and randomly extracted color patches, \emph{street image} SVHN (\textbf{sv})~\cite{netzer2011reading}, Synthetic Digits (\textbf{sy})~\cite{ganin2015unsupervised} generated from Windows fonts by various conditions, and \emph{handwritten} USPS (\textbf{up})~\cite{hull1994database}. Usually, 25,000 images are sampled for training and 9,000 for testing in \textbf{mt}, \textbf{mm}, \textbf{sv}, and \textbf{sy}. The entire 9,298 images in \textbf{up} are selected.

\textbf{Object classification.} Office-31~\cite{saenko2010adapting} contains 4,110 images in 31 categories collected from office environments in 3 domains: Amazon (\textbf{A}) with 2,817 images downloaded from amazon.com, Webcam (\textbf{W}) and DSLR (\textbf{D}) with 795 and 498 images taken by web camera and digital SLR camera with different photographical settings.

Office-Caltech~\cite{gong2013connecting} consists of the 10 overlapping categories shared by Office-31~\cite{saenko2010adapting} and Caltech-256 (\textbf{C})~\cite{griffin2007caltech}. Totally there are 2,533 images.

Office-Home~\cite{venkateswara2017deep} consists of about 15,500 images from 65 categories of everyday objects in office and home settings. There are 4 different domains: Artistic images (\textbf{Ar}), Clip Art (\textbf{Cl}), Product images (\textbf{Pr}) and Real-World images (\textbf{Rw}).

ImageCLEF, originated from ImageCLEF 2014 domain adaptation challenge\footnote{\url{http://imageclef.org/2014/adaptation}\label{ImageCLEF}}, consists of 12 object categories shared by ImageNet ILSVRC 2012 (\textbf{I}), Pascal VOC 2012 (\textbf{P}), and \textbf{C}. Totally there are 600 images for each domain with 50 for each category.

PACS~\cite{li2017deeper} contains 9,991 images of 7 object categories extracted from 4 different domains: Photo (\textbf{P}), Art paintings (\textbf{A}), Cartoon
(\textbf{C}) and Sketch (\textbf{S}). 

DomainNet~\cite{peng2019moment}, the largest DA dataset to date for object classification, contains about 600K images from 6 domains: Clipart, Infograph, Painting, Quickdraw, Real, and Sketch. There are totally 345 object categories.

\textbf{Sentiment classification of images.} SentiImage~\cite{lin2020multi} is a DA dataset with 4 domains for binary sentiment classification of images: \textit{social} Flickr and Instagram (\textbf{FI})~\cite{you2016building}, \textit{artistic} ArtPhoto (\textbf{AP})~\cite{machajdik2010affective}, \textit{social} Twitter I (\textbf{TI})~\cite{you2015robust}, and \textit{social} Twitter II (\textbf{TII})~\cite{borth2013large}. There are 23,308, 806, 1,269, and 603 images in these 4 domains, respectively.

\textbf{Vehicle counting.} WebCamT~\cite{zhang2017understanding} is a vehicle counting dataset from large-scale city camera videos with low resolution, low frame rate, and high occlusion. Totally there are 60,000 frames with vehicle bounding box and count annotations. For MDA, 8 cameras located in different intersections are selected, each with  more than 2,000 labeled images. We can view each camera as a domain.

\begin{figure*}[!t]
\begin{center}
\centering \includegraphics[width=1.0\linewidth]{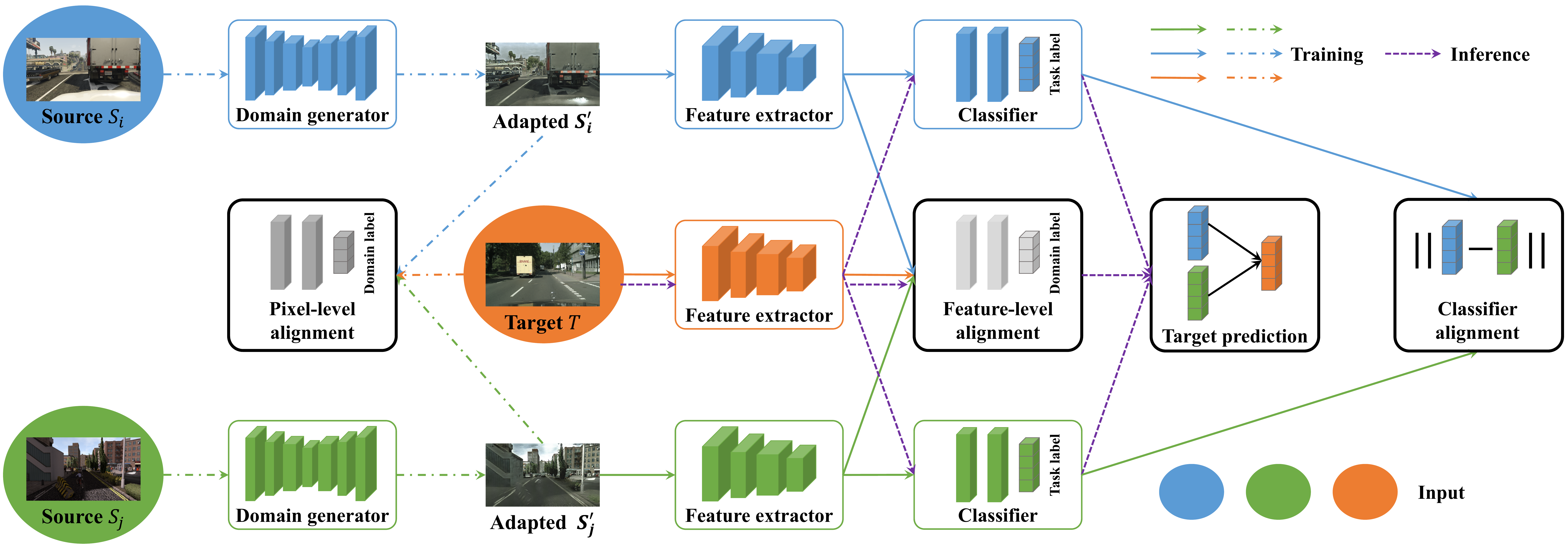}
\caption{Illustration of widely employed framework for MDA. The solid arrows and dashed dot arrows indicate the training of latent space transformation and intermediate domain generation, respectively. The dashed arrows represent the reference process. Most existing MDA methods can be obtained by employing different component details, enforcing some constraints, or slightly changing the architecture. Best viewed in color.}
\label{fig:Framework}
\end{center}
\end{figure*}

\textbf{Scene segmentation.} Sim2RealSeg contains 2 synthetic datasets (GTA, SYNTHIA) and 2 real datasets (Cityscapes, BDDS) for segmentation.  Cityscapes (CS)~\cite{cordts2016cityscapes} contains vehicle-centric urban street images collected from a moving vehicle in 50 cities from Germany and neighboring countries. There are 5,000 images with pixel-wise annotations into 19 classes. BDDS~\cite{yu2018bdd100k} contains 10,000 real-world dash cam video frames with a compatible label space with Cityscapes. GTA~\cite{richter2016playing} is a  vehicle-egocentric image dataset collected in the high-fidelity rendered computer game GTA-V. It contains 24,966 images (video frames) with 19 classes as Cityscapes. SYNTHIA~\cite{ros2016synthia} is a large synthetic dataset. To pair with Cityscapes, a subset, named SYNTHIA-RANDCITYSCAPES, is designed with 9,400 images which are automatically annotated with 16 compatible Cityscapes classes, one void class, and some unnamed classes. The common 16 classes are used for MDA.

\textbf{Sentiment classification of natural languages.} Amazon Reviews~\cite{chen2012marginalized} is a dataset of reviews on four kinds of products: Books (\textbf{B}), DVDs (\textbf{D}), Electronics (\textbf{E}), and Kitchen appliances (\textbf{K}). Reviews are encoded as 5,000 dimensional feature vectors of unigrams and bigrams and are labeled with binary sentiment. Each source has 2,000
labeled examples, and the target test set has 3,000 to 6,000 examples. 

Media Reviews~\cite{liu2017adversarial} contains 16 domains of product reviews and movie reviews for binary sentiment classification. 5 domains with 6,897 labeled samples are usually employed for MDA, including Apparel, Baby, Books, Camera taken from Amazon and MR from Rotten Tomato.

\textbf{Part-of-speech tagging.} The SANCL dataset~\cite{petrov2012overview} contains part-of-speech tagging annotations in 5 web domains: Emails (\textbf{E}), Weblogs (\textbf{W}), Answers (\textbf{A}), Newsgroups (\textbf{N}), and Reviews (\textbf{R}). 750 sentences from each source are used for training.

Unless otherwise specified, each domain is selected as the target and the rest domains are considered as the sources.  For WebCamT, 2 domains are randomly selected as the target. For Sim2RealSeg, MDA is often performed using the simulation-to-real setting~\cite{zhao2019multi}, \textit{i.e.} from synthetic GTA, SYNTHIA to real Cityscapes, BDDS. For SANCL, \textbf{N}, \textbf{R}, and \textbf{A} are used as target domains, while \textbf{E} and \textbf{W} are used as target domains~\cite{guo2018multi}.

\begin{table*}[!t]
\centering\scriptsize
\resizebox{\textwidth}{!}{%
\begin{tabular}
{ccccccccccc}
\specialrule{1pt}{1pt}{1pt}
\multirow{2}{*}{\textbf{Reference}} & \textbf{Feature} & \textbf{Feature} & \textbf{Feature} & \textbf{Feature} & \textbf{Classifier} & \multirow{2}{*}{\textbf{\#C}} & \textbf{Classifier} & \textbf{Task} & \multirow{2}{*}{\textbf{Dataset}} & \multirow{2}{*}{\textbf{Result}} \\
& \textbf{extractor} & \textbf{alignment method} & \textbf{alignment loss} & \textbf{alignment domains} & \textbf{alignment} & & \textbf{weight} & \textbf{backbone} & & \\
\hline
\cite{mancini2018boosting} & shared & --- & --- & --- & CT loss &  1 & --- & AlexNet & O, OC, P & 83.6, 91.8, 85.3 \\
\cite{guo2018multi} & shared & discrepancy & MMD & target and each source & --- & $M$ & PoS metric & AlexNet & AR, S & 84.8, 90.1 \\
\cite{hoffman2018algorithms} & shared & discrepancy & R\'enyi-divergence & target and each source & CT loss & 1 & --- & AlexNet & O & 87.6 \\
\cite{zhu2019aligning} & shared & discrepancy & MMD & target and each source & $\mathcal{L}1$ loss & $M$ & uniform & ResNet-50 & O, OH, IC & 90.2, 89.4, 74.1 \\
\cite{rakshit2019unsupervised} & unshared & discrepancy & $\mathcal{L}2$ distance & pairwise all domains & CT loss & 1 & --- & ResNet-50 & O, OC, IC & 88.3, 97.5, 91.2 \\
\hline
\multirow{3}{*}{\cite{peng2019moment}} & \multirow{3}{*}{shared} & \multirow{3}{*}{discrepancy} & \multirow{3}{*}{moment distance} & \multirow{3}{*}{pairwise all domains} & \multirow{3}{*}{$\mathcal{L}1$ loss} & \multirow{3}{*}{$M$} & \multirow{3}{*}{relative error} & LeNet-5 & D & 87.7 \\
 & & & & & & & &  ResNet-101 & OC & 96.4 \\
 & & & & & & & &  ResNet-101 & DN & 42.6\\
 \hline
\cite{guo2020multi} & shared & discrepancy & mixture distance & target and each source & CT loss & 1 & --- & BiLSTM & MR & 79.3 \\

\specialrule{1pt}{1pt}{1pt}

\cite{xu2018deep} & shared & discriminator & GAN loss & target and each source & --- & $M$ & perplexity score & AlexNet & D, O, IC & 74.2, 83.8, 80.8 \\
\cite{li2018extracting} & shared & discriminator & Wasserstein & pairwise all domains & CT loss & 1 & --- & AlexNet & D & 79.9 \\
\hline
\multirow{3}{*}{\cite{zhao2018adversarial}} & \multirow{3}{*}{shared} & \multirow{3}{*}{discriminator} & \multirow{3}{*}{$\mathcal{H}$-divergence} & \multirow{3}{*}{target and each source} & \multirow{3}{*}{CT loss} & \multirow{3}{*}{1} & \multirow{3}{*}{---} & BiLSTM & AR & 82.7 \\
& & & & & & & & AlexNet & D & 76.6\\
& & & & & & & & FCN8s & W & 1.4\\
\hline
\multirow{2}{*}{\cite{wang2019tmda}} & \multirow{2}{*}{shared} & \multirow{2}{*}{discriminator} & \multirow{2}{*}{Wasserstein} & \multirow{2}{*}{pairwise all domains} & \multirow{2}{*}{CT loss} & \multirow{2}{*}{1} & \multirow{2}{*}{---} & BiLSTM & AR & 84.5 \\
& & & & & & & & AlexNet & D & 83.4 \\
\hline
\multirow{2}{*}{\cite{zhao2020multi}} & \multirow{2}{*}{unshared} & \multirow{2}{*}{discriminator} & \multirow{2}{*}{Wasserstein} & \multirow{2}{*}{target and each source} & \multirow{2}{*}{---} & \multirow{2}{*}{$M$} & \multirow{2}{*}{Wasserstein} & LeNet-5 & D& 88.1 \\
& & & & & & & & AlexNet & O & 84.2\\
\specialrule{1pt}{1pt}{1pt}
\end{tabular}
}
\caption{Comparison of different latent space transformation methods for MDA, where `\#C', `CT loss', and `MMD' are short for the number of classifiers during reference ($M$ is the number of source domains), combined task loss, and maximum mean discrepancy, respectively. `Result' is the average performance of all target domains measured by accuracy for classification and counting error for vehicle counting.
}
\label{tab:LatentSpace}
\end{table*}

\section{Deep Multi-source Domain Adaptation}
\label{sec:MDA}

Existing methods on deep MDA primarily focus on the unsupervised, homogeneous, closed set, strongly supervised, one target, and target data available settings. That is, there is one target domain, the target data is unlabeled but available during the training process, the source data is fully labeled, the source and target data are observed in the same data space, and the label sets of all sources and the target are the same. In this paper, we focus on MDA methods under these settings.


There are some theoretical analysis to support existing MDA algorithms. Most theories are based on the seminal theoretical model~\cite{blitzer2008learning,ben2010theory}. \citeauthor{mansour2009mda}~\shortcite{mansour2009mda} assumed that the target distribution can be approximated by a mixture of the $M$ source distributions. Therefore, weighted combination of source classifiers has been widely employed for MDA. Moreover, tighter cross domain generalization bound and more accurate measurements on domain discrepancy can provide intuitions to derive effective MDA algorithms. \citeauthor{hoffman2018algorithms}~\shortcite{hoffman2018algorithms} derived a novel bound using DC-programming and calculated more accurate combination weights.
\citeauthor{zhao2018adversarial}~\shortcite{zhao2018adversarial} extended the generalization bound of seminal theoretical model to multiple sources under both classification and regression settings. Besides the domain discrepancy between the target and each source~\cite{hoffman2018algorithms,zhao2018adversarial}, \citeauthor{li2018extracting}~\shortcite{li2018extracting} also considered the relationship between pairwise sources and derived a tighter bound on weighted multi-source discrepancy. Based on this bound, more relevant source domains can be picked out.


Typically, some task models (\textit{e.g.} classifiers) are learned based on the labeled source data with corresponding task loss, such as cross-entropy loss for classification. Meanwhile, specific alignments among the source and target domains are conducted to bridge the domain shift so that the learned task models can be better transferred to the target domain. Based on the different alignment strategies, we can classify MDA into different categories. \emph{Latent space transformation} tries to align the latent space (\textit{e.g.} features) of different domains based on optimizing the discrepancy loss or adversarial loss. \emph{Intermediate domain generation} explicitly generates an intermediate adapted domain for each source that is indistinguishable from the target domain. The task models are then trained on the adapted domain. Figure~\ref{fig:Framework} summarizes the common overall framework of existing MDA methods.

\subsection{Latent Space Transformation}
\label{ssec:SpaceTransformation}

The two common methods for aligning the latent spaces of different domains are discrepancy-based methods and adversarial methods. We discuss these two methods below, and Table~\ref{tab:LatentSpace} summarizes key examples of each method.


\textbf{Discrepancy-based methods} explicitly measure the discrepancy of the latent spaces (typically features) from different domains by optimizing some specific discrepancy losses, such as maximum mean discrepancy (MMD)~\cite{guo2018multi,zhu2019aligning}, R\'enyi-divergence~\cite{hoffman2018algorithms}, $\mathcal{L}2$ distance~\cite{rakshit2019unsupervised}, and moment distance~\cite{peng2019moment}. \citeauthor{guo2020multi}~\shortcite{guo2020multi} claimed that different discrepancies or distances can only provide specific estimates
of domain similarities and that each distance has its pathological cases. Therefore, they consider the mixture of several distances~\cite{guo2020multi}, including $\mathcal{L}2$ distance, Cosine distance, MMD, Fisher linear discriminant, and Correlation alignment. Minimizing the discrepancy to align the features among the source and target domains does not introduce any new parameters that must be learned.

\textbf{Adversarial methods} try to align the features by making them indistinguishable to a discriminator. Some representative optimized objectives include GAN loss~\cite{xu2018deep}, $\mathcal{H}$-divergence~\cite{zhao2018adversarial}, Wasserstein distance~\cite{li2018extracting,wang2019tmda,zhao2020multi}. These methods aim to confuse the discriminator's ability to distinguishing whether the features from multiple sources were drawn from the same distribution. Compared with GAN loss and $\mathcal{H}$-divergence, Wasserstein distance can provide
more stable gradients even when the target and source distributions do not overlap~\cite{zhao2020multi}. The discriminator is often implemented as a network, which leads to new parameters that must be learned.

There are many modular implementation details for both types of methods, such as how to align the target and multiple sources, whether the feature extractors are shared, how to select the more relevant sources, and how to combine the multiple predictions from different classifiers.

\textbf{Alignment domains.} There are different ways to align the target and multiple sources. The most common method is to pairwise align the target with each source~\cite{xu2018deep,guo2018multi,zhao2018adversarial,hoffman2018algorithms,zhu2019aligning,zhao2020multi,guo2020multi}. Since domain shift also exists among different sources, several methods enforce pairwise alignment between every domain in both the source and target domains~\cite{li2018extracting,rakshit2019unsupervised,peng2019moment,wang2019tmda}.

\begin{table*}[!t]
\centering\scriptsize
\resizebox{\textwidth}{!}{%
\begin{tabular}
{ccccccccccc}
\specialrule{1pt}{1pt}{1pt}
\multirow{2}{*}{\textbf{Reference}} & \textbf{Domain} & \textbf{Pixel} &  \textbf{Feature} & \textbf{Feature} & \multirow{2}{*}{\textbf{\#C}} & \textbf{Classifier} & \textbf{Task} & \multirow{2}{*}{\textbf{Dataset}} & \multirow{2}{*}{\textbf{Task}} & \multirow{2}{*}{\textbf{Result}} \\
& \textbf{generator} & \textbf{alignment domains} & \textbf{alignment loss} & \textbf{alignment domains} & & \textbf{weight} & \textbf{backbone} & & & \\
\hline
    \multirow{2}{*}{\cite{russo2019towards}} & CoGAN  & \multirow{2}{*}{target and each source} & \multirow{2}{*}{GAN loss} & \multirow{2}{*}{target and each source} & \multirow{2}{*}{$M$} & \multirow{2}{*}{uniform} & \multirow{2}{*}{DeepLabV2} & \multirow{2}{*}{S2R-CS} & \multirow{2}{*}{seg} & \multirow{2}{*}{42.8} \\
& shared & & & & & & & & & \\
\hline
\multirow{2}{*}{\cite{zhao2019multi}} & CycleGAN  & \multirow{2}{*}{target and aggregated source}  & \multirow{2}{*}{GAN loss} & \multirow{2}{*}{target and each source} & \multirow{2}{*}{1} & \multirow{2}{*}{---} & \multirow{2}{*}{FCN8s} & S2R-CS  & \multirow{2}{*}{seg} & 41.4 \\
& shared & & & &  & & & S2R-BDDS & & 36.3\\
\hline
\multirow{2}{*}{\cite{lin2020multi}} & VAE+CycleGAN  & \multirow{2}{*}{target and combined source} & \multirow{2}{*}{---} & \multirow{2}{*}{---}  & \multirow{2}{*}{1} & \multirow{2}{*}{---} & \multirow{2}{*}{ResNet-18} & \multirow{2}{*}{SI} & \multirow{2}{*}{cls} & \multirow{2}{*}{68.1} \\
& unshared & & & & & &  & & & \\
\specialrule{1pt}{1pt}{1pt}
\end{tabular}
}
\caption{Comparison of different intermediate domain generation methods for MDA, where `\#C', `seg', and `cls' are short for the number of classifiers during reference ($M$ is the number of source domains), segmentation, and classification, respectively. `Result' is the average performance of all target domains measured by accuracy for classification and mean intersection-over-union (mIoU) for segmentation.}
\label{tab:IntermediateDomain}
\end{table*}

\textbf{Weight sharing of feature extractor.} Most methods employ shared feature extractors to learn domain-invariant features. However, domain invariance may be detrimental to discriminative power. On the contrary, \citeauthor{rakshit2019unsupervised}~\shortcite{rakshit2019unsupervised} adopted one feature extractor for each source and target pair with unshared weights, while \citeauthor{zhao2020multi}~\shortcite{zhao2020multi} first pre-trained one feature extractor for each source and then mapped the target into the feature space of each source. Correspondingly, there are $M$ and $2M$ feature extractors. Although unshared feature extractors can better align the target and sources in the latent space, this substantially increases the number of parameters in the model.

\textbf{Classifier alignment.} Intuitively, the classifiers trained on different sources may result in misaligned predictions for the target samples that are close to the domain boundary. By minimizing specific classifier discrepancy, such as $\mathcal{L}$1 loss~\cite{zhu2019aligning,peng2019moment}, the classifiers are better aligned, which can learn a generalized classification boundary for target samples mentioned above. Instead of explicitly training one classifier for each source, many methods focus on training a compound classifier based on specific combined task loss, such as normalized activations~\cite{mancini2018boosting} and bandit controller~\cite{guo2020multi}.


\textbf{Target prediction.} After aligning the features of target and source domains in the latent space, the classifiers trained based on the labeled source samples can be used to predict the labels of a target sample. Since there are multiple sources, it is possible that they will yield different target predictions. One way to reconcile these different predictions is to uniformly average the predictions from different source classifiers~\cite{zhu2019aligning}. However, different sources may have different relationships with the target, \textit{e.g.} one source might better align with the target, so a non-uniform, weighted averaging of the predictions leads to better results. Weighting strategies, known as a \emph{source selection process}, include uniform weight~\cite{zhu2019aligning}, perplexity score based on adversarial loss~\cite{xu2018deep}, point-to-set (PoS) metric using Mahalanobis distance~\cite{guo2018multi}, relative error based on source-only accuracy~\cite{peng2019moment}, and Wasserstein distance based weights~\cite{zhao2020multi}.

Besides the source importance, \citeauthor{zhao2020multi}~\shortcite{zhao2020multi} also considered the sample importance, \textit{i.e.} different samples from the same source may still have different similarities from the target samples. The source samples that are closer to the target are distilled (based on a manually selected Wasserstein distance threshold) to fine-tune the source classifiers. Automatically and adaptively selecting the most relevant training samples for each source remains an open research problem.

\subsection{Intermediate Domain Generation}
\label{ssec:DomainGeneration}

Feature-level alignment only aligns high-level information, which is insufficient for fine-grained predictions, such as pixel-wise 
semantic segmentation~\cite{zhao2019multi}. Generating an intermediate adapted domain with pixel-level alignment, typically via GANs~\cite{goodfellow2014generative}, can help address this problem. 

\textbf{Domain generator.} Since the original GAN is highly under-constrained, some improved versions are employed, such as Coupled GAN (CoGAN) in~\cite{russo2019towards} and CycleGAN in MADAN~\cite{zhao2019multi}. Instead of directly taking the original source data as input to the generator~\cite{russo2019towards,zhao2019multi}, \citeauthor{lin2020multi}~\shortcite{lin2020multi} used a variational autoencoder to map all source and target domains to a latent space and then generated an adapted domain from the latent space. \citeauthor{russo2019towards}~\shortcite{russo2019towards} then tried to align the target and each adapted domain, while \citeauthor{lin2020multi}~\shortcite{lin2020multi} aligned the target and combined adapted domain from the latent space.  \citeauthor{zhao2019multi}~\shortcite{zhao2019multi} proposed to aggregate different adapted domains using a sub-domain aggregation discriminator and cross-domain cycle discriminator, where the pixel-level alignment is then conducted between the aggregated and target domains. \citeauthor{zhao2019multi}~\shortcite{zhao2019multi} and \citeauthor{lin2020multi}~\shortcite{lin2020multi} showed that the semantics might change in the intermediate representation, and that enforcing a semantic consistency before and after generation can help preserve the labels.

\textbf{Feature alignment and target prediction.} Feature-level alignment is often jointly considered with pixel-level alignment. Both alignments are usually achieved by minimizing the GAN loss with a discriminator. One classifier is trained on each adapted domain~\cite{russo2019towards} and the multiple predictions for a given target sample are averaged. Only one classifier is trained on the aggregated domain~\cite{zhao2020multi} or on the combined adapted domain~\cite{lin2020multi} which is obtained by a unique generator from the latent space for all source domains. The comparison of these methods are summarized in Table~\ref{tab:IntermediateDomain}.

\section{Conclusion and Future Directions}
\label{sec:Conclusion}
In this paper, we provided a survey of recent MDA developments in the deep learning era. We motivated MDA, defined different MDA strategies, and summarized the datasets that are commonly used for performing MDA evaluation. Our survey focused on a typical MDA setting, \textit{i.e.} unsupervised, homogeneous, closed set, and one target MDA. We classified these methods into different categories, and compared the representative ones technically and experimentally. 
We conclude with several open research directions:

\textbf{Specific MDA strategy implementation.} As introduced in Section~\ref{sec:Definition}, there are many types of MDA strategies, and implementing an MDA strategy according to the specific problem requirement would likely yield better results than a one-size-fits-all MDA approach. Further investigation is needed to determine which MDA strategies work the best for which types of problems. Also, real-world applications may have a small amount of labeled target data; determining how to include this data and what fraction of this data is needed for a certain performance remains an open question.


\textbf{Multi-modal MDA.} The labeled source data may be of different modalities, such as LiDAR, radar, and image. Further research is needed to find techniques for fusing different data modalities in MDA. A further extension of this idea is to have varied modalities in defferent sources as well as partially labeled, multi-modal sources.

\textbf{Incremental and online MDA.} Designing incremental and online MDA algorithms remains largely unexplored and may provide great benefit for real-world scenarios, such as updating deployed MDA models when new source or target data becomes available. 



\clearpage
\tiny\bibliographystyle{named}
\bibliography{MDA_Survey}

\end{document}